\documentclass[conference]{IEEEtran}

\usepackage[T1]{fontenc}
\usepackage[utf8]{inputenc}

\usepackage{amsmath,amssymb}
\usepackage{booktabs}
\usepackage{multirow}
\usepackage{array}

\usepackage{graphicx}
\usepackage{xcolor}

\usepackage{enumitem}

\usepackage{url}
\usepackage{hyperref}

\usepackage{balance}

\hypersetup{
  colorlinks=true,
  linkcolor=blue,
  citecolor=blue,
  urlcolor=blue
}

\setlist[itemize]{leftmargin=*, topsep=2pt, itemsep=2pt, parsep=0pt}
\setlist[enumerate]{leftmargin=*, topsep=2pt, itemsep=2pt, parsep=0pt}

\title{Translational Gaps in Graph Transformers for Longitudinal EHR Prediction:\\
A Critical Appraisal of GT-BEHRT}

\author{
\IEEEauthorblockN{Krish Tadigotla}
\IEEEauthorblockA{
Texas Academy of Mathematics and Science (TAMS)\\
University of North Texas\\
Email: \href{mailto:KrishTadigotla@my.unt.edu}{KrishTadigotla@my.unt.edu}
}
}

\makeatletter
\renewenvironment{abstract}{%
  \par\noindent\normalfont\textnormal{Abstract---}\ignorespaces
}{\par}

\makeatother

\begin{document}
\maketitle

\begin{abstract}
Background: Transformer-based architectures have advanced predictive modeling on longitudinal electronic health records (EHRs) through self-supervised pretraining. However, many implementations represent clinical encounters as unordered code sets, potentially discarding diagnostically relevant intra-visit relational structure. Graph--transformer hybrids propose to address this limitation by explicitly encoding visit-level relationships while preserving long-range temporal dependencies.

Objective: To critically appraise GT-BEHRT, a graph--transformer framework evaluated on MIMIC-IV intensive care outcomes and All of Us Research Program heart failure prediction, assessing whether reported performance gains translate to clinically actionable risk stratification under contemporary standards for medical AI evaluation.

Methods: We conducted systematic appraisal across seven evaluation dimensions: representation fidelity, pretraining comparability, cohort construction transparency, metric sufficiency (discrimination versus calibration), fairness auditing rigor, reproducibility infrastructure, and deployment feasibility. Assessment criteria were anchored to TRIPOD guidelines for prediction models and emerging machine learning fairness frameworks \cite{tripod,hardt2016equality}.

Results: GT-BEHRT achieved AUROC $94.37 \pm 0.20$, AUPRC $73.96 \pm 0.83$, and F1 $64.70 \pm 0.85$ on All of Us heart failure prediction within 365 days \cite{poulain2024gtbehrt}. Despite strong discrimination, six translational gaps were identified: (i) calibration analysis absent (no calibration curves, Brier score, or expected calibration error), (ii) fairness auditing incomplete (no Equal Opportunity/Equalized Odds with uncertainty), (iii) selection bias risk from cohort restrictions, (iv) limited phenotype and prediction-horizon sensitivity analyses, (v) decision-analytic utility not assessed, and (vi) deployment feasibility insufficiently characterized for clinical settings.

Conclusion: GT-BEHRT is a meaningful architectural advance in EHR representation learning, yet the evidence base supporting clinical deployment remains incomplete. Calibration-aware evaluation, formal fairness auditing with uncertainty quantification, transparent cohort reporting with sensitivity analyses, decision-analytic benchmarking, and deployment-oriented evaluation are prerequisites before graph--transformer EHR systems can credibly inform clinical decision-making.
\end{abstract}

\begin{IEEEkeywords}
Electronic Health Records, Graph Transformers, Clinical Prediction Models, Calibration, Algorithmic Fairness, All of Us Research Program
\end{IEEEkeywords}

\section{Introduction}
Heart failure (HF) remains a major driver of morbidity and rehospitalization, with 30-day all-cause readmissions commonly exceeding 20\% in large U.S.\ administrative cohorts \cite{hf_readmission}. Earlier identification of individuals at elevated risk for HF events could enable targeted preventive interventions, optimize resource allocation, and reduce preventable hospitalizations. EHR systems capture longitudinal patient trajectories---diagnoses, medications, procedures, and laboratory values---that may encode preclinical progression markers. However, EHR data present substantial challenges: temporal irregularity (visits occur at unpredictable intervals), structural heterogeneity (variable code counts per encounter and implicit relationships among co-prescribed medications and concurrent diagnoses), and systematic measurement biases (healthcare access patterns, coding practice variation, and informative missingness).

Methodologically, deep learning for EHR prediction has evolved through distinct phases. Early recurrent neural network (RNN) architectures with attention mechanisms modeled temporal dependencies but faced optimization challenges on long sequences and offered limited parallelizability \cite{ma2017dipole}. Transformer encoders addressed these constraints through self-attention and enabled large-scale self-supervised pretraining via masked-code prediction objectives. Representative implementations include BEHRT and CEHR-BERT, which tokenize clinical codes and incorporate temporal embeddings \cite{li2019behrt,pang2021cehrbert}. Despite architectural sophistication, many transformer EHR models represent encounters as unordered code sets---implicitly requiring attention layers to infer intra-visit structure (e.g., medication--diagnosis couplings and procedure--indication relationships) from co-occurrence alone.

Graph-based EHR methods provide an alternative by explicitly encoding medical concepts and their relationships as graph-structured objects. Hypergraph approaches aim to capture higher-order interactions within encounters \cite{xu2023hypehr}. However, graph-centric architectures have historically struggled to efficiently model long-range temporal dependencies and lack standardized self-supervised pretraining paradigms comparable to transformer-based masked-objective frameworks.

Hybrid graph--transformer designs attempt to reconcile these strengths. GT-BEHRT exemplifies this approach: each clinical visit is represented as a graph with medical codes as nodes, a graph transformer produces a visit embedding capturing intra-visit structure, and a BERT-style temporal transformer models the sequence of visit embeddings to encode longitudinal dynamics \cite{poulain2024gtbehrt}. Evaluated on the All of Us Research Program---a large, demographically diverse cohort---GT-BEHRT reported strong discrimination for heart failure prediction within 365 days \cite{poulain2024gtbehrt}.

While superior discrimination (AUROC, AUPRC) is necessary for predictive utility, it is insufficient for clinical translation. This review addresses a fundamental question: does the current evidence base support the translational claim that improved representational capacity yields clinically actionable and equitably performant risk prediction? We evaluate GT-BEHRT against contemporary standards for medical AI, including calibration requirements, fairness auditing expectations, cohort transparency norms, and deployment feasibility criteria.

\section{Methods of Critical Appraisal}
We conducted systematic critical appraisal of GT-BEHRT using a seven-dimension framework adapted from TRIPOD guidelines and emerging machine learning fairness standards \cite{tripod,hardt2016equality}:
\begin{enumerate}
\item \textbf{Representation Design:} Visit tokenization versus graph-based encoding; alignment of inductive biases with clinical workflow; multimodal integration capacity.
\item \textbf{Pretraining Strategy:} Objective alignment with downstream tasks; compute-normalized comparability; ablation rigor and causal attribution strength.
\item \textbf{Cohort Construction:} Inclusion/exclusion transparency; selection bias risk; censoring logic and window definitions; attrition reporting completeness.
\item \textbf{Metric Sufficiency:} Discrimination versus calibration and clinical utility; threshold selection justification; uncertainty quantification practices.
\item \textbf{Fairness Auditing:} Reporting of Equal Opportunity/Equalized Odds; subgroup-specific calibration; confidence intervals for subgroup estimates; bias mitigation documentation.
\item \textbf{Reproducibility Infrastructure:} Code availability; hyperparameter completeness; environment specification; independent verification feasibility.
\item \textbf{Deployment Feasibility:} Latency and memory footprint reporting; robustness to missingness and site heterogeneity; drift monitoring and auditability; workflow integration pathways.
\end{enumerate}

Quantitative comparisons were anchored to GT-BEHRT's published results for MIMIC-IV benchmarks and All of Us heart failure prediction \cite{poulain2024gtbehrt}. Contextual assessment incorporated recent EHR foundation models, generative architectures, and hypergraph-based prediction systems \cite{yang2023transformehr,li2023hibehrt,rasmy2021medbert,xu2023hypehr}.

\section{GT-BEHRT Architecture and Contributions}
\subsection{Visit-as-Graph Representational Schema}
GT-BEHRT operationalized each encounter as a graph wherein nodes correspond to medical codes (e.g., diagnoses, medications, procedures), supplemented by a virtual visit node for aggregation \cite{poulain2024gtbehrt}. This directly targets a limitation of tokenization approaches: codes within an encounter lack inherent order, and positional encodings on unordered sets are a weak proxy for genuine relational structure.

\subsection{Hierarchical Two-Level Transformer Factorization}
The model factorized representation learning into: (i) a graph transformer that processes each visit graph to produce a visit embedding, and (ii) a temporal transformer encoder that operates over visit embeddings to model longitudinal dynamics \cite{poulain2024gtbehrt}. This hierarchical design reduces effective sequence length from individual codes to aggregated visits, mitigating quadratic attention costs and potentially improving scalability.

\subsection{Dual-Stage Self-Supervised Pretraining}
GT-BEHRT used two-phase pretraining: Stage~1 masked reconstruction at the node level, and Stage~2 sequence-level objectives including missing-node prediction and visit-type prediction \cite{poulain2024gtbehrt}. Ablations suggested both stages contributed to downstream performance. However, baseline models differ in training recipes and compute budgets, complicating causal attribution of gains strictly to architecture.

\section{Quantitative Performance Comparison}
Table~\ref{tab:quant} reports GT-BEHRT's performance on MIMIC-IV mortality prediction and All of Us heart failure prediction, alongside representative baselines reproduced from the GT-BEHRT experimental table \cite{poulain2024gtbehrt}. Metrics are mean $\pm$ standard deviation across five random seeds.

\noindent\textbf{Critical contextual note:} F1 scores were computed using a fixed threshold of 0.5; the authors noted this may be suboptimal when predicted probabilities are not calibrated around 0.5 \cite{poulain2024gtbehrt}. For imbalanced outcomes and varying prevalence across deployment sites, threshold-agnostic metrics and clinically justified operating points are preferable.

\begin{table*}[t]
\centering
\caption{Quantitative comparison on GT-BEHRT reported benchmarks (mean $\pm$ std across five runs). Metrics are AUROC/AUPRC/F1. Baseline results are reproduced from GT-BEHRT.}
\label{tab:quant}
\renewcommand{\arraystretch}{1.15}
\setlength{\tabcolsep}{5.2pt}
\begin{tabular}{lccc|ccc}
\toprule
\multirow{2}{*}{\textbf{Model}} &
\multicolumn{3}{c|}{\textbf{MIMIC-IV Mortality}} &
\multicolumn{3}{c}{\textbf{All of Us HF (365d)}} \\
& \textbf{AUROC} & \textbf{AUPRC} & \textbf{F1} & \textbf{AUROC} & \textbf{AUPRC} & \textbf{F1} \\
\midrule
Dipole \cite{ma2017dipole} &
$92.24\pm0.68$ & $59.56\pm1.20$ & $56.96\pm1.77$ &
$86.87\pm1.91$ & $60.26\pm2.32$ & $55.34\pm1.86$ \\
HiTANet \cite{luo2020hitanet} &
$92.67\pm0.32$ & $62.46\pm1.44$ & $57.63\pm2.37$ &
$91.59\pm1.07$ & $65.29\pm1.66$ & $59.19\pm1.41$ \\
BEHRT \cite{li2019behrt} &
$92.67\pm0.19$ & $60.54\pm1.66$ & $55.76\pm2.87$ &
$87.07\pm1.76$ & $61.92\pm1.64$ & $54.81\pm1.31$ \\
CEHR-BERT \cite{pang2021cehrbert} &
$93.31\pm0.23$ & $64.79\pm2.37$ & $59.22\pm2.43$ &
$87.37\pm1.74$ & $63.68\pm2.82$ & $57.28\pm2.10$ \\
\midrule
\textbf{GT-BEHRT} \cite{poulain2024gtbehrt} &
$\mathbf{94.11\pm0.31}$ & $\mathbf{65.55\pm1.65}$ & $\mathbf{64.07\pm2.13}$ &
$\mathbf{94.37\pm0.20}$ & $\mathbf{73.96\pm0.83}$ & $\mathbf{64.70\pm0.85}$ \\
\bottomrule
\end{tabular}
\end{table*}

\section{Comparative Landscape of EHR Modeling Paradigms}
Table~\ref{tab:compare} summarizes representative longitudinal EHR modeling families relevant to evaluating graph--transformer hybrids. To avoid repetitive critique, each paradigm's limitation is stated distinctly.

\begin{table*}[t]
\centering
\caption{Representative longitudinal EHR modeling paradigms relevant to graph--transformer evaluation.}
\label{tab:compare}
\renewcommand{\arraystretch}{1.12}
\setlength{\tabcolsep}{4.6pt}
\begin{tabular}{p{2.8cm} p{2.4cm} p{3.0cm} p{7.0cm}}
\toprule
\textbf{Study} & \textbf{Paradigm} & \textbf{Core Representation} & \textbf{Distinct Limitation / Tension} \\
\midrule
Dipole \cite{ma2017dipole} & RNN + attention & Visit sequence (bag of codes) & Harder scaling to long histories; limited parallelism; weaker compatibility with modern masked pretraining. \\
\midrule
BEHRT \cite{li2019behrt} & Transformer & Code tokens + age/segment embeddings & Flattens intra-visit structure; long sequences remain costly; visit boundaries encoded indirectly. \\
\midrule
CEHR-BERT \cite{pang2021cehrbert} & Transformer + SSL & Time-aware embeddings over EHR tokens & Improves temporal encoding but still treats encounters as weakly structured; irregular sampling remains challenging. \\
\midrule
Med-BERT \cite{rasmy2021medbert} & Transformer pretraining & Structured diagnosis sequences & Diagnosis-centric pretraining; limited encounter-level relational structure and multimodal integration. \\
\midrule
Hi-BEHRT \cite{li2023hibehrt} & Hierarchical transformer & Hierarchical aggregation for long EHRs & Long-sequence handling via compression; choices can reduce interpretability and increase compute complexity. \\
\midrule
TransformEHR \cite{yang2023transformehr} & Encoder--decoder generative transformer & Generative longitudinal EHR modeling & Generative objectives may not align with classification utility; evaluation can overemphasize proxy metrics versus decision benefit. \\
\midrule
HypEHR \cite{xu2023hypehr} & Hypergraph model & Set-level hyperedges among codes & Graph construction and hyperedge design can dominate results; temporal modeling may be weaker than dedicated temporal transformers. \\
\midrule
GT-BEHRT \cite{poulain2024gtbehrt} & Graph transformer + temporal transformer & Visit graphs $\rightarrow$ visit embeddings $\rightarrow$ temporal encoder & Strong discrimination, but translation remains uncertain without calibration, formal fairness auditing, decision-curve utility, and deployment-oriented evaluation. \\
\bottomrule
\end{tabular}
\end{table*}

\section{Critical Analysis: Six Translational Gaps}
Each gap follows: problem $\rightarrow$ clinical impact $\rightarrow$ actionable recommendation.

\subsection{Gap 1: Calibration Analysis Entirely Absent}
\textbf{Problem:} GT-BEHRT reported discrimination metrics (AUROC, AUPRC) and threshold-specific F1 scores but provided no calibration assessment---no calibration curves, Brier score, expected calibration error, or subgroup-calibration analyses \cite{poulain2024gtbehrt}.

\textbf{Clinical impact:} In clinical risk prediction, calibrated probabilities are often more decision-relevant than rank-order discrimination. A model with high AUROC that overestimates risk can increase false alarms and overtreatment; underestimation can delay care. Subgroup miscalibration can systematically disadvantage particular patient populations.

\textbf{Actionable recommendation:} Report calibration curves with 95\% confidence bands; compute Brier score with decomposition; report expected calibration error; evaluate subgroup calibration; assess post-hoc calibration under temporal validation and quantify effects on decision outcomes.

\subsection{Gap 2: Fairness Auditing Incomplete Under Contemporary Standards}
\textbf{Problem:} Subgroup-stratified performance was reported, but formal fairness metrics (Equal Opportunity / Equalized Odds) and uncertainty quantification were not provided; subgroup calibration was not evaluated \cite{poulain2024gtbehrt,hardt2016equality}.

\textbf{Clinical impact:} Disparities in sensitivity or false-positive rates can exacerbate inequities---for example, delayed diagnosis in underrepresented groups or disproportionate downstream workup burden. Without confidence intervals, the statistical significance of observed disparities remains unclear.

\textbf{Actionable recommendation:} Report Equal Opportunity and Equalized Odds with bootstrap 95\% confidence intervals; test disparity significance; evaluate subgroup calibration and standardized net benefit parity at clinically meaningful thresholds; document mitigation strategies and quantify fairness--accuracy tradeoffs.

\subsection{Gap 3: Selection Bias From Cohort Restrictions Not Fully Characterized}
\textbf{Problem:} The model excluded patients with fewer than two visits; exclusion magnitude and differences between excluded and included participants were not fully foregrounded with sensitivity analyses \cite{poulain2024gtbehrt}.

\textbf{Clinical impact:} Evaluating primarily higher-utilization patients can inflate performance and reduce transportability to sparse-history patients and care-fragmented populations, which often represent higher-risk vulnerable groups.

\textbf{Actionable recommendation:} Provide attrition diagrams; compare baseline characteristics of excluded vs.\ included groups; vary minimum-visit thresholds and report performance by visit-count bins; test sparse-history robustness through stratified analyses or augmentation.

\subsection{Gap 4: Phenotype and Prediction-Horizon Sensitivity Underexplored}
\textbf{Problem:} HF prediction within 365 days is clinically plausible, but phenotyping and horizon choice strongly influence task definition; systematic sensitivity analyses were not emphasized in the main evidence narrative \cite{poulain2024gtbehrt}.

\textbf{Clinical impact:} Models can appear robust under one operational definition but fail under alternative phenotypes or clinically preferred horizons, limiting reproducibility and stakeholder confidence in model generalization.

\textbf{Actionable recommendation:} Vary phenotype code sets and case definitions; evaluate multiple horizons and report horizon-stratified metrics with confidence intervals; assess stability across rolling time windows; surface phenotype specifications prominently in methods.

\subsection{Gap 5: Clinical Decision Utility Not Demonstrated}
\textbf{Problem:} Improved discrimination was not linked to decision-making benefit. No decision-curve analysis quantified net benefit versus treat-all/treat-none baselines; threshold selection lacked clinician-validated justification \cite{poulain2024gtbehrt}.

\textbf{Clinical impact:} AUROC gains may not translate to improved outcomes if net benefit is unchanged at clinically relevant thresholds or if workflow constraints make recommended actions impractical.

\textbf{Actionable recommendation:} Conduct decision-curve analysis; report net benefit and standardized net benefit at clinically meaningful thresholds; compare to established risk tools; validate thresholds with clinicians; pursue prospective evaluation to quantify workflow impact.

\subsection{Gap 6: Deployment Feasibility Constraints Insufficiently Characterized for Clinical Use}
\textbf{Problem:} GT-BEHRT reported partial runtime statistics (e.g., inference time per 1{,}000 patients on an NVIDIA T4) \cite{poulain2024gtbehrt}, but did not establish clinical-grade feasibility: end-to-end latency including graph construction and data retrieval, memory footprint under typical hospital infrastructure, and robustness under real-world missingness, EHR backfill, site heterogeneity, and drift were not evaluated as deployment outcomes.

\textbf{Clinical impact:} Clinical decision support often requires low-latency inference and reliable performance under evolving data pipelines. Without end-to-end deployment benchmarks and robustness testing, operational failure modes remain unknown, creating barriers to safe clinical adoption.

\textbf{Actionable recommendation:} Report end-to-end latency (including preprocessing and graph construction) and memory on commodity clinical hardware; benchmark performance--cost frontiers under compute constraints; stress-test missingness and site shifts; develop drift monitoring and retraining criteria; explore distillation, quantization, and caching to reduce computational burden.

\section{Reproducibility Infrastructure}
GT-BEHRT's public code release is a strength \cite{gtbehrt_github}. Nonetheless, reproducibility gaps remain that merit emphasis.

Cohort definition transparency is critical: phenotyping lists, inclusion--exclusion logic, censoring rules, and window definitions should be foregrounded to enable independent verification. Executable cohort scripts, including workbench-compatible notebooks and SQL code for preprocessing, facilitate reproducible cohort reconstruction and auditing when data access permits.

Training specification completeness directly influences compute-normalized comparability. Complete hyperparameter configurations, training duration, learning rate schedules, and reported compute budgets enable fair comparison against contemporaneous baselines.

\noindent\textbf{Actionable recommendation:} Release cohort scripts where feasible, include attrition diagrams with exact counts, provide machine-readable hyperparameters and compute budgets in supplementary materials, and publish synthetic validation artifacts or summary statistics when data access is restricted.

\section{Clinical Translation Feasibility}
Graph--transformer hybrids introduce deployment barriers requiring explicit evaluation to meet contemporary standards for clinical AI systems.

\subsection{Computational and Operational Complexity}
Graph construction requires robust engineering for heterogeneous codes, missingness, and longitudinal updates. Multi-stage inference pipelines can increase latency and maintenance burden as coding systems evolve. Error propagation through graph construction to final predictions can degrade reliability if not carefully managed.

\subsection{Governance and Auditability Requirements}
Clinical use requires audit trails and governance infrastructure. Attention visualizations alone may be insufficient for explanation; drift monitoring is essential for multi-site deployment to detect distribution shifts and trigger retraining.

\noindent\textbf{Actionable recommendation:} Benchmark end-to-end latency on commodity clinical hardware; characterize failure modes under missing data and coding variation; define drift detection and retraining policies; evaluate explanation methods with clinicians to ensure actionability; pilot shadow-mode deployment prior to prospective trials; maintain comprehensive audit logs linking predictions to training versions and data cohorts.

\section{Future Research Priorities}
We prioritize near-term directions with high translational impact and feasibility within current data and computational infrastructure.
\begin{enumerate}
\item \textbf{Calibration-first evaluation:} Report calibration curves overall and by demographic subgroup, compute Brier score decomposition, report expected calibration error across deciles, and conduct temporal calibration validation.
\item \textbf{Comprehensive fairness auditing:} Measure Equal Opportunity and Equalized Odds with bootstrap confidence intervals; stratify calibration assessment by demographic groups; compute standardized net benefit parity; document fairness--accuracy tradeoffs.
\item \textbf{Transparent cohort reporting:} Publish attrition diagrams with exact counts; compare baseline characteristics of excluded versus included groups; vary minimum-visit thresholds and report performance stratified by visit frequency; document missing-data patterns and imputation strategies.
\item \textbf{Decision-analytic utility:} Conduct decision-curve analysis across clinically meaningful decision thresholds; compare net benefit to existing risk tools; validate threshold recommendations through clinician engagement; pursue prospective evaluation to quantify outcomes impact.
\item \textbf{Deployment-oriented benchmarking:} Report end-to-end latency and memory on representative clinical hardware; conduct compute-normalized comparisons; stress-test robustness to missingness patterns, site shifts, and temporal drift; develop monitoring infrastructure for real-world tracking.
\end{enumerate}

\section{Conclusion}
GT-BEHRT advances EHR representation learning by explicitly encoding intra-visit relational structure via visit graphs and modeling longitudinal dynamics with a temporal transformer, achieving strong discrimination for tasks including All of Us heart failure prediction \cite{poulain2024gtbehrt}. However, architectural innovation has outpaced evidentiary rigor for clinical translation. Calibration assessment, formal fairness auditing with uncertainty quantification, transparent cohort reporting with sensitivity analyses, decision-analytic utility quantification, and deployment feasibility evaluation must be treated as first-class research outcomes alongside discrimination metrics. Until these gaps are addressed through deployment-oriented evaluation and prospective validation, GT-BEHRT and similar systems remain research prototypes rather than clinically actionable tools ready for integration into real-world care delivery.

\balance
\bibliographystyle{IEEEtran}
\bibliography{references}

\end{document}